\documentclass{article}

\usepackage{times}
\usepackage{graphicx} 
\usepackage{subfigure} 

\usepackage{natbib}

\usepackage{algorithm}
\usepackage{algorithmic}

\usepackage{amssymb}
\usepackage{amsmath}
\usepackage{amsfonts}
\usepackage{latexsym}
\usepackage{booktabs}
\usepackage{siunitx}

\usepackage{hyperref}

\newcommand{\PP}{\mathbb{P}}

\newcommand{\by}{\mathbf{y}}
\newcommand{\bx}{\mathbf{x}}

\newcommand{\bz}{\mathbf{z}}

\usepackage[accepted]{deepstruct2017}

\icmltitlerunning{Low-Rank Hidden State Embeddings for Viterbi Sequence Labeling}

\begin{document} 

\twocolumn[
\icmltitle{Low-Rank Hidden State Embeddings for Viterbi Sequence Labeling}

\icmlsetsymbol{equal}{*}

\begin{icmlauthorlist}
\icmlauthor{Dung Thai}{umass}
\icmlauthor{Shikhar Murty}{umass}
\icmlauthor{Trapit Bansal}{umass}
\icmlauthor{Luke Vilnis}{umass}
\icmlauthor{David Belanger}{umass}
\icmlauthor{Andrew McCallum}{umass}
\end{icmlauthorlist}

\icmlaffiliation{umass}{CICS, University of Massachusetts, Amherst, USA}

\icmlcorrespondingauthor{Dung Thai}{dthai@cs.umass.edu}

\icmlkeywords{sequence tagging, latent dynamic CRF}

\vskip 0.3in
]

\printAffiliationsAndNotice{}  

\begin{abstract}

In textual information extraction and other sequence labeling tasks it is now common to use recurrent neural networks (such as LSTM) to form rich embedded representations of long-term input co-occurrence patterns.  Representation of output co-occurrence patterns is typically limited to a hand-designed graphical model, such as a linear-chain CRF representing short-term Markov dependencies among successive labels.  This paper presents a method that learns embedded representations of latent output structure in sequence data.  Our model takes the form of a finite-state machine with a large number of latent states per label (a latent variable CRF), where the state-transition matrix is factorized---effectively forming an embedded representation of state-transitions capable of enforcing long-term label dependencies, while supporting exact Viterbi inference over output labels. We demonstrate accuracy improvements and interpretable latent structure in a synthetic but complex task based on CoNLL named entity recognition.

\end{abstract}

\section{Introduction}

\label{sec:intro}

Neural networks have long been used for prediction tasks involving complex structured outputs \cite{lecun2006tutorial,collobert11,DBLP:journals/corr/LampleBSKD16}. In structured prediction, output variables obey local and global constraints that are difficult to satisify using purely local feedforward prediction from an input representation. For example, in sequence tagging tasks such as named entity recognition, the outputs must obey several hard constraints e.g., I-PER cannot follow B-ORG. The results of \cite{collobert11} show a significant improvement when such structural output constraints are enforced by incorporating a linear-chain graphical model that captures the interactions between adjacent output variables. The addition of a graphical model to enforce output consistency is now common practice in deep structured prediction models for tasks such as sequence tagging \citep{DBLP:journals/corr/LampleBSKD16} and image segmentation \citep{chen2014semantic}.

From a probabilistic perspective, the potentials of a probabilistic graphical model over the output variables $y$ are often parameterized using a deep neural network that learns global features of the input $x$ \cite{lecun2006tutorial,collobert11,DBLP:journals/corr/LampleBSKD16}. This approach takes advantage of deep architectures to learn robust feature representations for $x$, but is limited to relatively simple pre-existing graphical model structures to model the interactions among $y$. 

This paper presents work in which feature learning is used not only to learn rich representations of inputs, but also to learn latent output structure. We present a model for sequence tagging that takes the form of a latent-variable conditional random field \citep{quattoni07,sutton07,morency07}, where interactions in the latent state space are parametrized by low-rank embeddings. This low-rank structure allows us to use a larger number of latent states learning rich and interpretable substructures in the output space without overfitting. Additionally, unlike LSTMs, the model permits exact MAP and marginal inference via the Viterbi and forward-backward algorithms. Because the model learns large numbers of latent hidden states, interactions among $y$ are not limited to simple Markov dependencies among labels as in most deep learning approaches to sequence tagging.

Previous work on representation learning for structured outputs has taken several forms. Output-embedding models such as \cite{srikumar2014learning} have focused on learning low-rank similarity among label vectors $y$, with no additional latent structure. The input-output HMM \citep{bengio95} incorporates learned latent variables, parameterized by a neural network, but the lack of low-rank structure limits the size of the latent space. 
Structured prediction energy networks \citep{belanger2016structured} use deep neural networks to learn global output representations, but do not allow for exact inference and are difficult to apply in cases when the number of outputs varies independently of the number of inputs, such as entity extraction systems.

In this preliminary work, we demonstrate the utility of learning a large embedded latent output space on a synthetic task based on CoNLL named entity recognition (NER). We consider the task synthetic because\comment{we condition the output model on} we employ input features involving only single tokens, which allows us to better examine the effects of both learned latent output variables and low-rank embedding structure. (The use of NER data is preferable, however, to completely synthetically generated data because its real-world text naturally contains easily interpretable complex latent structure.) We demonstrate significant accuracy gains from low-rank embeddings of large numbers of latent variables in output space, and explore the interpretable latent structure learned by the model. These results show promise for future application of low-rank latent embeddings to sequence modeling tasks involving more complex long-term memory, such as citation extraction, resum{\' e}s, and semantic role labeling.

\section{Related Work}
\label{sec:related}
The ability of neural networks to efficiently represent local context features sometimes allows them to make surprisingly good independent decisions for each structured output variable \cite{collobert11}. However, these independent classifiers are often insufficient for structured prediction tasks where there are strong dependencies between the output labels \cite{collobert11,DBLP:journals/corr/LampleBSKD16}. A natural solution is to use these neural feature representations to parameterize the factors of a conditional random field \cite{lafferty01} for joint inference over output variables \cite{collobert11,jaderberg14,DBLP:journals/corr/LampleBSKD16}. However, most previous work restricts the linear-chain CRF states to be the labels themselves---learning no additional output structure.

The latent dynamic conditional random field (LDCRF) learns additional output structure beyond the labels by employing hidden states (latent variables) with Markov dependencies, each associated with a label; it has been applied to human gesture recognition \cite{morency07}.  The dynamic conditional random field (DCRF) learns a factorized representation of each state \cite{sutton07}.  The hidden-state conditional random field (HCRF) also employs a Markov sequence of latent variables, but the latent variables are used to predict a single label rather than a sequence of labels; it has been applied to phoneme recognition \cite{gunawardana05} and gesture recognition \cite{quattoni07}.  All these models learn output representations while preserving the ability to perform exact joint inference by belief propagation.  While the above use a log-linear parameterization of the potentials over latent variables, the input-output HMM \citep{bengio95} uses a separate neural network for each source state to produce transition probabilities to its destination states.

Experiments in all of the above parameterizations use only a small hidden state space due to the large numbers of parameters required.  In this paper we enable a large number of states by using a low-rank factorization of the transition potentials between latent states, effectively learning distributed embeddings for the states. This is superficially similar to the label embedding model of \cite{srikumar2014learning}, but that work learns embeddings only to model similarity between observable output labels, and does not learn a latent output state structure.

\section{Embedded Latent CRF Model}
\label{sec:model}

We consider the task of sequence labeling: given an input sequence $\textbf{x} = \{x_1, x_2, \ldots, x_T\}$, find the corresponding output labels $\textbf{y} = \{y_1, y_2, \ldots, y_T\}$ where each output $y_i$ is one of $N$ possible output labels. Each input $x_i$ is associated with a feature vector $f_i \in \mathbb{R}^n$, such as that produced by a feed-forward or recurrent neural network.

The models we consider will associate each input sequence with a sequence of hidden states $\{z_1, z_2, \ldots, z_T\}$. These discrete hidden states capture rich transition dynamics of the output labels. We consider the case where the number of hidden states $M$ is much larger than the number of output labels, $M >> N$.

Given the above notation, the energy for a particular configuration is:
\begin{align}
\mathcal{E}(\by, \bz| \bx) = \sum_{t=1}^T ( & \psi_{zf}(f_t, z_t) + \psi_{zy}(z_t, y_t) \nonumber \\ &+ \psi_{zz}(z_t, z_{t+1})) \label{eq:energy}
\end{align}
where $\psi$ are scalar scoring functions of their arguments. 
$\psi_{zf}(f_t, z_t)$ and $\psi_{zy}(z_t, y_t)$ are the local scores for the interaction between the input features and the hidden states, and the hidden state and the output state, respectively. $\psi_{zz}(z_t, z_{t+1})$ are the scores for transitioning from a hidden state $z_t$ to hidden state $z_{t+1}$.
The distribution over output labels is given by:
\begin{align}
\PP(\by|\bx) = \frac{1}{Z} \sum_{\bz} \exp \left( \mathcal{E}(\by, \bz| \bx) \right); \label{eq:py}
\end{align}
$Z = \sum_{\by} \sum_{\bz} \exp \left( \mathcal{E}(\by, \bz| \bx) \right)$ is the partition function.

In the case of our Embedded Latent CRF model, as in the LDCRF model, latent states are deterministically partitioned to correspond to output values. That is, the number of latent states is a multiple of the number of output values, and $\psi_{zy}=0$ for pairs in the partitioning and $-\infty$ otherwise. The other scoring functions are learned as global bilinear parameter matrices.

In order to manage large numbers of latent states without overfitting, the Embedded Latent CRF enforces an additional restriction that the scoring function $\psi_{zz}$ possess a low-rank structure: that is, $\psi_{zz}(z_i,z_j)=z_i^\top U V^\top z_j$, where $U$ and $V$ are skinny rectangular matrices and $z_i,z_j$ are represented by one-hot vectors.

While inference in this model is tractable using tree belief propagation even when learning $\psi_{zy}$, the deterministic factors make it especially simple to implement.

Computing the quantities involved in \eqref{eq:py} can be carried out efficiently with dynamic programming using the forward algorithm, as in HMMs. To see this, note that to compute the numerator $\sum_{\bz} \exp \left( \mathcal{E}(\by, \bz| \bx) \right)$, given an output label $\by$, we can fold the local scores $\psi_{zf}$ and $\psi_{zy}$ \eqref{eq:energy} into one score $\psi_{zfy}(f_t, z_t; y_t)$, and summing the resulting energy corresponds exactly to the forward algorithm in a CRF with $M$ states.
The partition function can also be computed by dynamic programming:
\begin{align*}
Z &= \sum_{\by} \sum_{\bz} \exp \left( \mathcal{E}(\by, \bz| \bx) \right) \\
&= \sum_{\bz} \sum_{\by} \prod_t  \exp \left(  \psi_{zf}(z_t, f_t)   
  + \psi_{zz}(z_t, z_{t+1}) \right) \\ & \exp(\psi_{zy}(z_t, y_t))
\end{align*}

At test time we perform MAP inference using the exact Viterbi algorithm, which can be done as in the above dynamic program, replacing sums with maxes.

\section{Experiments}
\label{sec:experiments}

We demonstrate the benefits of latent and embedded large-cardinality state spaces on a synthetic task based on CoNLL-2003 named entity recognition \cite{DBLP:conf/conll/SangM03}, which provides easily interpretable inputs and outputs to explore.  With IOB encoding the dataset has eight output labels representing non-entities, as well as inside and beginning of person, location, organization and miscelaneous enties (B-PER is not needed). We consider the task synthetic because we use relatively impoverished input features to explore the capacity of the output representation. We use the BiLSTM+CRF featurization from \cite{DBLP:journals/corr/LampleBSKD16}, but instead of using BiLSTM, we produce local potentials from a feedforward network conditioned on word- and character-level features from the current time-step only. We demonstrate that performance of this model is improved by increasing the size of the latent state space (perhaps unsurprisingly), and that significant further improvement can be obtained from learning low-rank embeddings of the latent states. Qualitatively, we also show the latent states learn interpretable structure.

\paragraph{Training Details:} 
All models are implemented using TensorFlow. We use the hyperparameter settings from the LSTM+CRF model of \cite{DBLP:journals/corr/LampleBSKD16}, with the exception that we use minibatches of size 20 and the Adam optimizer \cite{DBLP:journals/corr/KingmaB14} with a learning rate of 1e-3 and $\epsilon$ of 1e-6. We initialise our word level embeddings using pretrained 100 dimensional skip-n-gram embeddings \cite{DBLP:conf/emnlp/LingTAFDBTL15} where available, and use Glorot initialisation \cite{DBLP:journals/jmlr/GlorotB10} otherwise. 

For the Embedded Latent CRF, we learn 16 dimensional embeddings (the matrices $U$ and $V$ have rank 16) for all but the 16 hidden state ELCRF, for which we learn 8 dimensional embeddings. We train all models for 200 epochs with early-stopping on the validation set. We use the train-dev-test split of the CoNLL-2003 dataset.

\subsection{Quantitative Results}

\textbf{Table \ref{tab:crf-res}} reports the performance of all the models. We see that the performance on this structured prediction task is improved both by increasing the size of the latent state space, and by embedding the states into a lower dimensional space (i.e. a low-rank factorization of the log-space transition potential). The benefits of a complex output space are especially important when using a restricted set of input features, as noted in \citep{liang2008structure}.

\begin{table}[h]
\centering
\small
\begin{tabular}{c|c|c} 
\hline
\abovespace\belowspace
    {\textbf{\#States}} & {\textbf{LDCRF F1}} & {\textbf{ELCRF F1}} \\ \midrule
    8 & 81.88 & - \\
    16 & 83.69 & 84.03 \\
    32 & 84.31 & 84.58 \\
    64 & 84.52 & 85.02 \\
    128 & 84.36 & 86.29 \\
    256 & \textbf{84.82} & 86.83 \\
    512 & 84.78 & \textbf{86.91} \\ \bottomrule
 \end{tabular}
\caption{NER field F1 on the CoNLL'03 test set with synthetically simple input representation.  Note that  learning an embedded state representation (ELCRF) improves accurcy over not learning an embedded representation (LDCRF).}\label{tab:crf-res}
\end{table}

\subsection{Qualitative Insight}
We now turn to a qualitative analysis of the ELCRF model. Table \ref{tab:ner} gives examples of some hidden states and tokens for which they were activated. First, we observe that the model has discovered separate latent states for surnames (228) and surname nobiliary particles (192).  The latter almost always transitions to a surname state, and capturing this special transition signature (as distinct from {\sc Per} label generically, as in a traditional CRF without latent states) improves accuracy when the surname is poorly associated with the {\sc Per} label.

We also observe the model's ability to detect phrase boundaries. This is true not only for {\sc Per} phrases where the model identifies boundaries by identifying first names and last names but also for {\sc Misc} phrases and {\sc Org} phrases. We observe that state 257 fires everytime an {\sc I-Misc} token is followed by an {\sc I-Misc} token---signalling the start of an {\sc I-Misc} phrase, and state 272 fires at the end of that phrase. Similarly, state 392 signals the start of an {\sc Org} phrase and 445 signals the end.

\begin{table}[!ht]
\centering
\small
\begin{tabular}{c|S} \toprule
    {\textbf{state}} & {\textbf{tokens}} \\ \midrule
    {192}		&  {Van, Dal, De, Manne, Jan, Den, Della, Der}  \\
    {204} & {Hendrix, Lien, Werner, Peter, Sylvie, Jack} \\
    {228}    &  {Miller, Cesar, Jensen, Dickson, Abbott} \\ 
    {283}   & {British, German, Polish, Australian} \\
    {297} & {911, 310, 150, 11} \\
    {269} & {Korean-related, Beijing-funded, Richmond-based} \\ \bottomrule   
\end{tabular}
\caption{Tokens predicted for hidden states}\label{tab:ner}
\end{table}

The latent state space also learns block structure in the state transition matrix which gives the joint prediction a long-term bidirectional ``memory,'' encouraging unusual and beneficial interpretation of other parts of the sequence.  For example, in the sentence ``Boston 's Mo Vaughn went 3-for-3 with a walk , stole home ...,'' our model with factorized latent states was correctly able to label ``Boston'' as {\sc I-Org} (the team) due to the longer range context of a baseball game, whereas the model without latent states incorrectly labels it {\sc I-Loc}.  In the phrase ``Association for Relations Across the Taiwan Straits'' our model correctly labels the entire sequence as an {\sc Org} using a special state for ``the,'' while the traditional model loses context at ``the'' and labels the last two words as a {\sc Loc}.

\section{Conclusion and Future Work}

We present a method for learning output representations in finite-state sequence modeling. Our Embedded Latent CRF learns state embeddings by representing the transition matrix in a large latent state space with low-rank factorization.  Unlike most recent work that learns input representations, but settles with simple Markov dependencies among the given output labels, our approach learns output representations of a large number of memory-providing, expressive and interpretable states, avoids overfitting due to factorization, and maintains tractable exact inference plus maximum likelihood learning using dynamic programming. In future work we will apply this model to non-synthetic sequence labeling tasks involving complex joint predictions and long term memory, such as citation extraction, r{\' e}sum{\' e} field extraction, and semantic role labeling.

{\small
\bibliography{example_paper}
\bibliographystyle{icml2017}
}

\end{document}